\documentclass{article}
\usepackage{spconf,amsmath,graphicx}
\usepackage{algorithmic}
\usepackage{algorithm}
\usepackage{hyperref}

\title{SupervisorBot: NLP-Annotated Real-Time Recommendations of Psychotherapy Treatment Strategies with Deep Reinforcement Learning}
\name{Baihan Lin$^1$, Guillermo Cecchi$^2$, Djallel Bouneffouf$^2$}
\address{
  $^1$Columbia University\\
  $^2$IBM Thomas J. Watson Research Center\\
{baihan.lin@columbia.edu, gcecchi@us.ibm.com, djallel.bouneffouf@ibm.com}}
\begin{document}
%\ninept
%
\maketitle
\begin{abstract}

We propose a recommendation system that suggests treatment strategies to a therapist during the psychotherapy session in real-time. Our system uses a turn-level rating mechanism that predicts the therapeutic outcome by computing a similarity score between the deep embedding of a scoring inventory, and the current sentence that the patient is speaking. The system automatically transcribes a continuous audio stream and separates it into turns of the patient and of the therapist and perform real-time inference of their therapeutic working alliance. The dialogue pairs along with their computed working alliance as ratings are then fed into a deep reinforcement learning recommendation system where the sessions are treated as users and the topics are treated as items. Other than evaluating the empirical advantages of the core components on an existing dataset of psychotherapy sessions, we demonstrate the effectiveness of this system in a web app.

\end{abstract}

\section{Introduction}

Mental illness is not only a severe healthcare problem in the US (1 in 5 estimated by National Institute of Mental Health) but also a major global issue \cite{patel2018lancet}. However, most countries including the states face severe shortage of mental health practioners, such as psychiatrists and clincal psychologists \cite{satiani2018projected}. In recent two years, this demand gap was especially amplified by the toll of COVID-19 pandemic on everyone's mental health \cite{wang2020investigating}. Current education systems and training programs cannot catch up to this trend because each licensed therapist requires years of continual learning and supervised training. Even when a therapist is ripe for independent practice, many still seek weekly supervision from ``supervisors'', who are usually a more senior therapist that have seen many more years of patients and serve as ``a crucial triad of learning difficulties that tend to confront beginning therapists in their training'' \cite{watkins2013being}. These supervisors provide necessary guidance and periodic feedback to junior therapists with respect to their development of mindedness, psychotherapist identities and treatment roadblocks they face in their own cases.

\begin{figure*}[tb]
\centering
\begin{tabular}{l  c  r}
    \includegraphics[width=.38\linewidth]{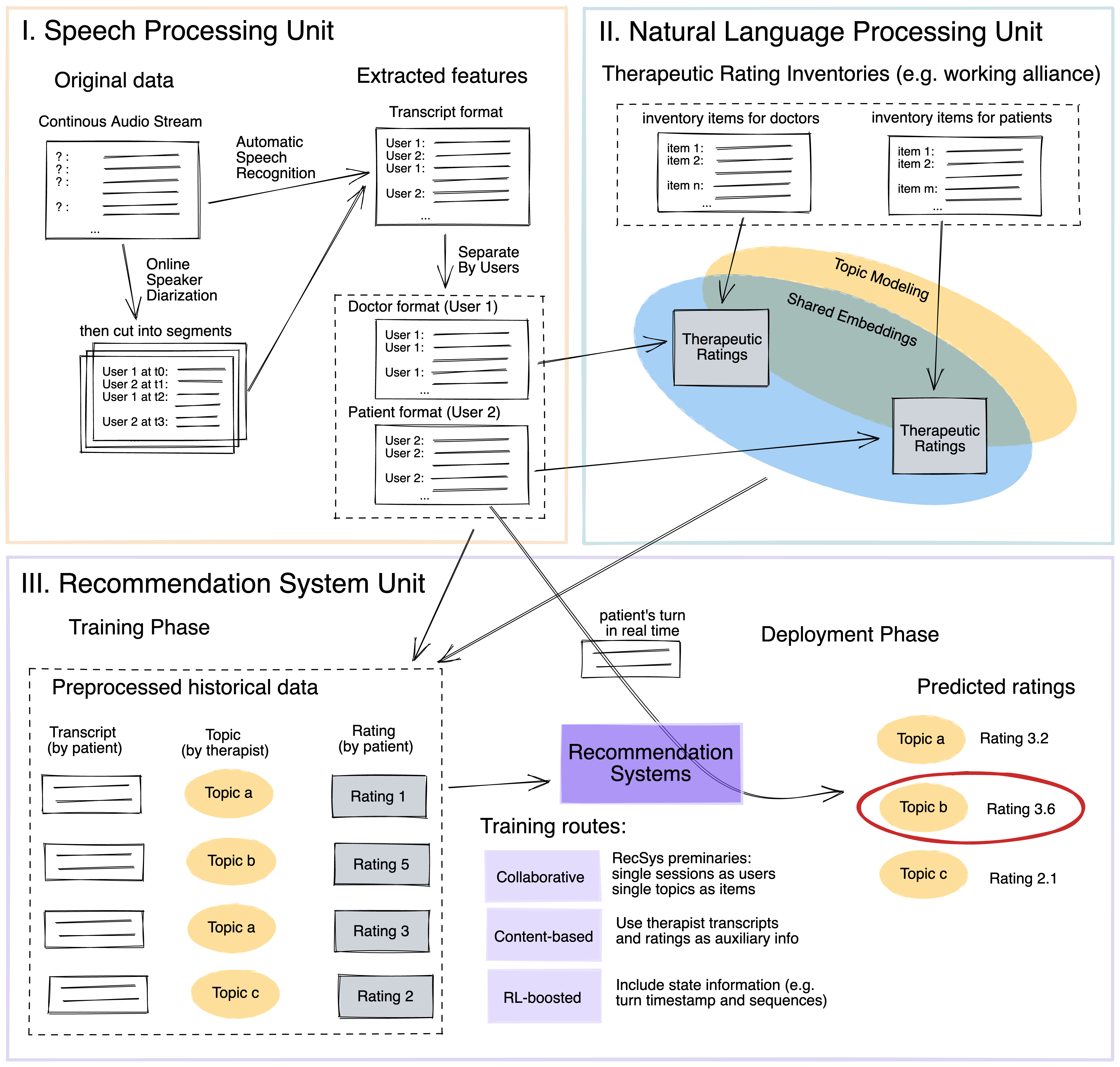} &     \includegraphics[width=.3\linewidth]{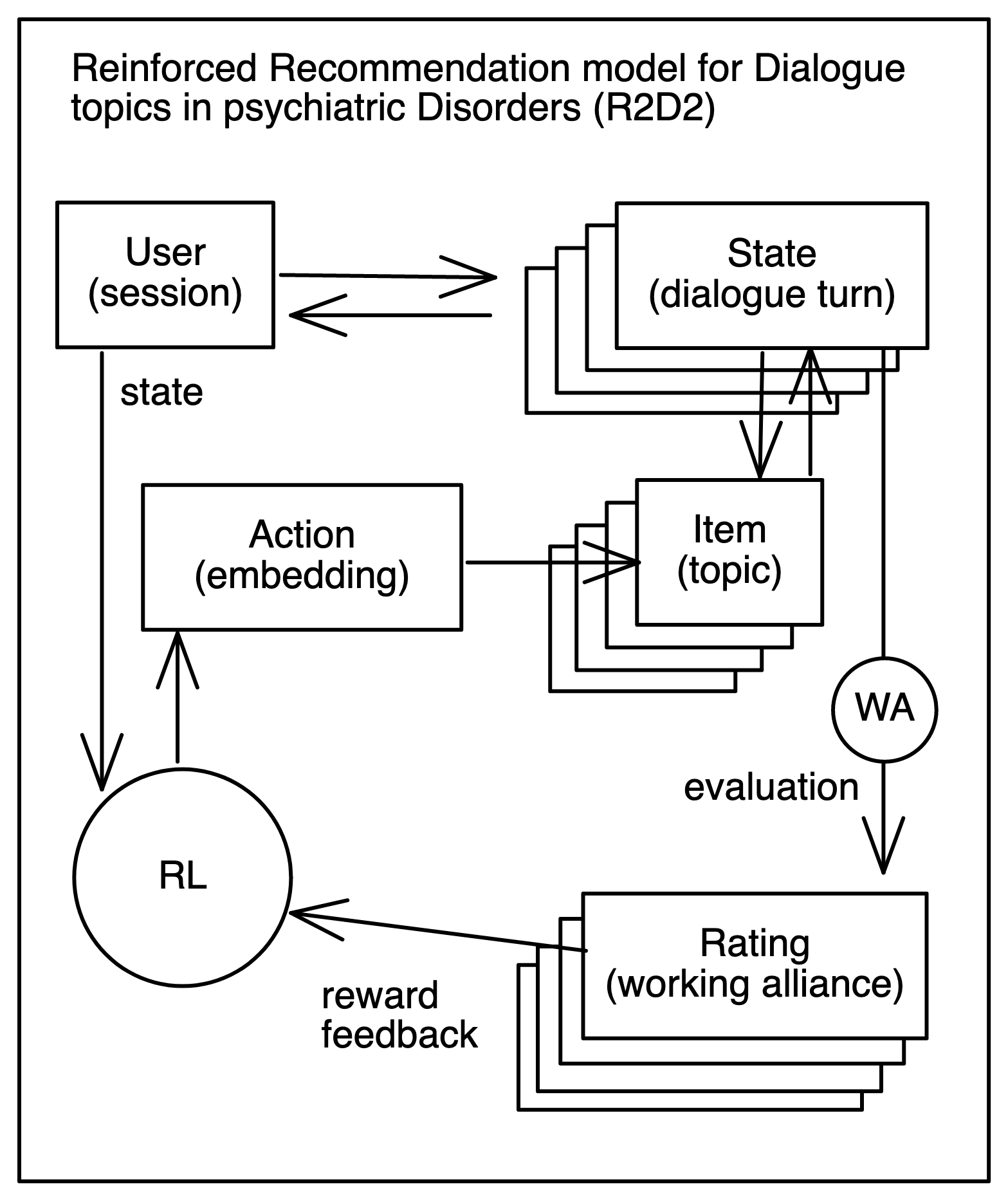} &
    \includegraphics[width=.27\linewidth]{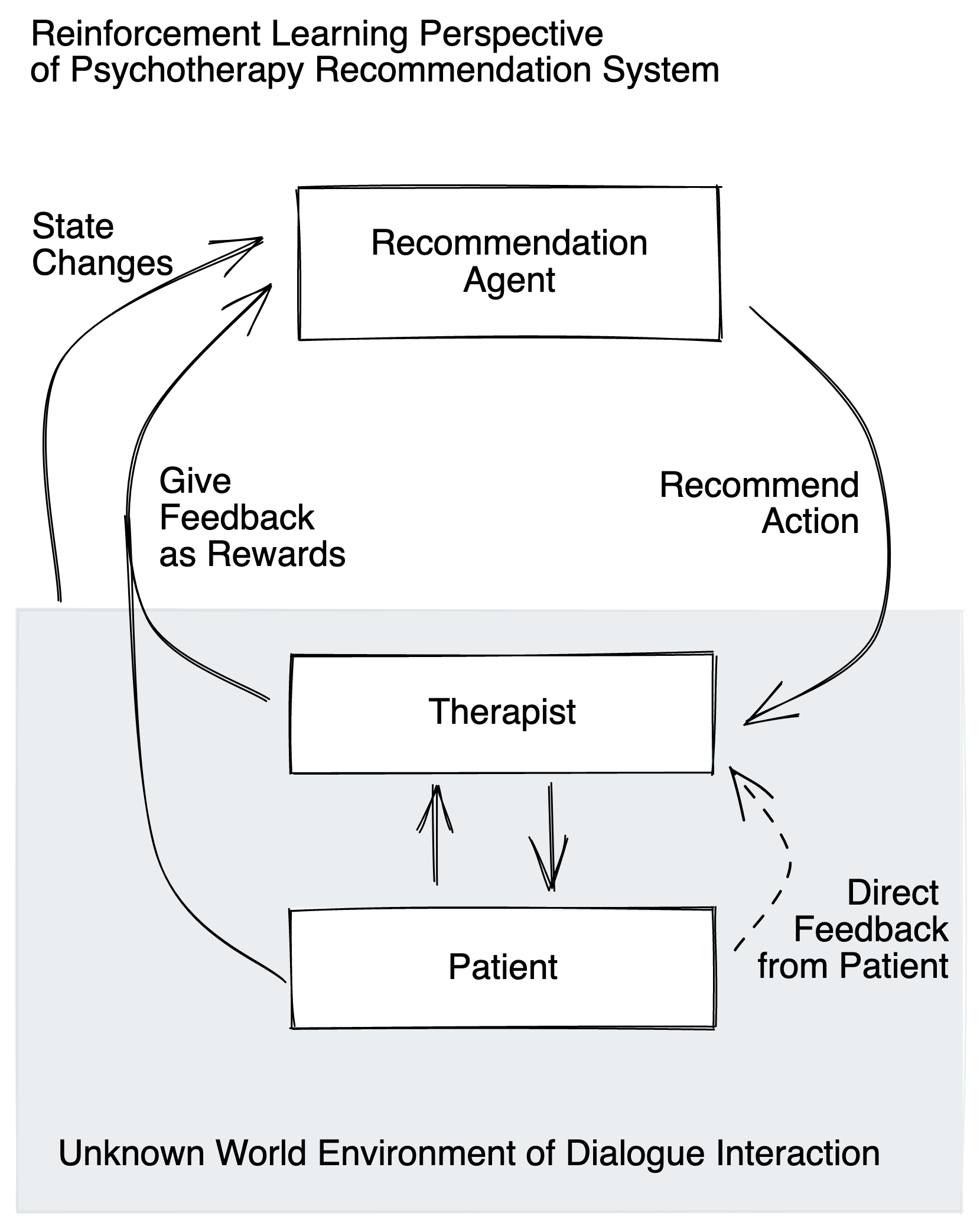} 
    \end{tabular}
\par\caption{Analytical Framework of SupervisorBot. (A) Major components of the system. (B) R2D2 model flowchart. (C) Reinforcement learning framework of the psychotherapy recommendation system problem.
}\label{fig:pipeline}
\end{figure*}

% \begin{figure}[tb]
% \centering
% % \begin{minipage}{.37\linewidth}
%   \includegraphics[width=\linewidth]{Figures/wai_keytable.png}
% \par\caption{Example inventory items and scales
% }\label{fig:wai}
% \end{figure}

% \begin{figure*}[tb]
% % \end{minipage}
% % \hfill
% % \begin{minipage}{.56\linewidth}
%     \includegraphics[width=\linewidth]{Figures/results.png}

% \caption{Empirical results on MiniVox and AlexStreet datasets
% }\label{fig:results}
% % \end{minipage}
% \end{figure*}

In this work, we present SupervisorBot, a virtual AI companion that provides real-time feedback and recommends treatment strategy to the therapists while they are conducting their own psychotherapy. Like a supervisor, SupervisorBot offers feedback and guidance that are case-dependent. Like a supervisor, SupervisorBot has seen thousands of historical therapy sessions and case studies.
% from experienced therapists. 
The base of our recommendation system relies on a rating system that evaluates how good a treatment strategy is. As the mental state of a patient can be complicated to characterize, we gravitates our approach towards well-defined clinical outcomes. The working alliance is such a psychological concept that is shown to be highly predictive of the success of psychotherapy in clinical setting \cite{Wampold2015}.
It describes several important cognitive and emotional components of the relationship between these two agents in conversation, including the agreement on the goals to be achieved and the tasks to be carried out, and the bond, trust and respect to be established over the course of the dialogue \cite{Bordin79}. 
% It measures the tendency for communication partners to align with each other both in their verbal and non-verbal behaviors.
% , and the strength of the alliance is one of the main factors that predict success in the clinical setting of psychotherapy \cite{Wampold2015}. 
% Traditionally, the alliance can only be measured through post hoc self-reports of the participants on very limited point-scales questionnaire valuation \cite{horvath1981exploratory}. This approach does not make use of the nuances afforded by natural language, is time-consuming and difficult to follow through systematically outside of research studies; even more so is the evaluation of individual dialogue turns over the course of each session.
% Instead, 
In \cite{lin2022deep}, we developed a natural language processing (NLP) approach to infer this quantity in real-time as ratings.
Here we propose the Reinforced Recommendation model for Dialogue topics in psychiatric Disorders (R2D2), a the first ever recommendation system of dialogue topics proposed for the psychotherapy setting. It transcribes the session in real-time, predicts the therapeutic outcome as a turn-level rating, and recommends treatment strategy that is best for the current context and state of the psychotherapy. 
It is the first step to solving the global issue of mental health by augmenting the treatment and education of clinical practitioners with a recommendation system of therapeutic strategy.

% It is web-based, interactive, informative, and learns continually. 
% There has not been such a system, as the one we are current demonstrating, that can (1) record and transcribe the session in real-time, (2) assess the quality of the conversation in a turn-level resolution based on the conversational alignment, and (3) recommend treatment strategy that is best for the current context and state of the psychotherapy. 
% Having a such tool would be very useful for practical applications such as training TV interviews, monitoring psychotherapy, assisting job assessments, and building better AI assistants to interact with general public.
% Our real-time AI system accomplishes all three objectives. 
% It is web-based, interactive, informative, and learns continually. 

% It is web-based,  
%  Here we present an approach to quantify the degree of patient-therapist alliance by projecting each turn in a therapeutic session onto the representation of clinically established working alliance inventories, using language modeling to encode both turns and inventories. This allows us not only to quantify the overall degree of alliance but also to identify granular patterns its dynamics over shorter and longer time scales. We also discuss how our approach may be used as a companion tool to provide feedback to the therapist and to augment learning opportunities for training therapists. 

\section{Methods}

% \subsection{Working alliance analysis}

Fig \ref{fig:pipeline} is an outline of the analytic framework. The continuous audio stream is fed into the system. First, we perform the speaker diarization (e.g. using real-time solutions such as \cite{lin2021speaker,lin2020voiceid,lin2020speaker}) which separates audio into dyads of doctor-patient, which are then transcribed into natural language turns for real-time downstream analyses.

\textbf{Therapeutic quality ratings.}
After we obtain a relatively well diarization result, we can configure the quality assessment setting by specifying a proper inventory. In this system, we use the Working Alliance Inventory (WAI), a set of self-report measurement questionnaire that quantifies the therapeutic bond, task agreement, and goal agreement \cite{horvath1981exploratory,tracey1989factor,martin2000relation}. 
% The WAI aims to (1) measure alliance factors across all types of therapy, (2) document the relationship between the alliance measure and the corresponding theoretical constructs underlying the measure, and (3) related the alliance measure to a unified theory of therapeutic change \cite{horvath1994working}. 
Operationally, our goal is to derive from these 36 items three alliance scales: the task scale, the bond scale and the goal scale. They measures the three major themes of psychotherapy outcomes: (1) the collaborative nature of the dialogue participants' relationship; (2) the affective bond between them, and (3) their capabilities to agree on treatment-related short-term tasks and long-term goals. The score corresponding to the three scales comes from a key table
% (Fig \ref{fig:wai}) 
which specifies the positivity or the sign weight to be applied on the questionnaire answer when summing in the end. 
% The full scale is simply the sum of the scores of the three scales.
% The key table is like a weighting matrix that specifies the directionalities of the scales. 

% \begin{figure}[tb]
% \centering
% \begin{minipage}{.37\linewidth}
%   \includegraphics[width=\linewidth]{Figures/wai_keytable.png}
% 
% \par\caption{Example inventory items and scales
% }\label{fig:wai}
% \end{minipage}
% \hfill
% \begin{minipage}{.56\linewidth}
%     \includegraphics[width=\linewidth]{Figures/results.png}
% 
% \caption{Empirical results on MiniVox and AlexStreet datasets
% }\label{fig:results}
% \end{minipage}

% \end{figure}

\textbf{Transcription and real-time rating assessment.}
Now we are ready for real-time quality annotation. Given the audio stream for a given user, we first transcribe the diarized audio stream with standard automatic speech recognition module \cite{adorf2013web}. 
% When we have the transcribed sentence and our inventory statements, we compare their deep embeddings.
% Algorithm \ref{alg:waa} outlines the process.Say, we have two dialogue participants, a patient and a therapist. We denote each patient response turn as $S^p_i$ followed by a therapist response turn $S^t_i$. They are treated as a dialogue pair. The inventories of working alliance questionnaires also come in pairs: $I^p$ for the patient (or client), and  $I^t$ for the therapist. They each consist of 36 statements. 
Following the approach proposed in \cite{lin2022deep,lin2022working,lin2022unsupervised,lin2022voice}, 
we embed both the dialogue turns and WAI items with deep sentence or paragraph embeddings (in this case, Doc2Vec \cite{le2014distributed}), and then compute the cosine similarity between the embedding vectors of the turn and its corresponding inventory vectors. With that, for each turn (either by patient or by therapist), we obtain a 36-dimension working alliance score, which we may save in a relational database as in \cite{lin2022knowledge}. 

\textbf{Topic modeling as recommendation items.}
First, we define the ``items'', ``users'', ``contents'' and ``ratings'' in our recommendation system. Here, the ``items'' the system recommends are treatment strategies. In this example, we represents these strategies as a topic that the therapist should initiate or continue for the next turn. Given a large text corpus of many psychotherapy sessions, as in \cite{lin2022neural} we can first perform topic modeling
% (e.g. \cite{miao2016neural,miao2017discovering,nan2019topic,dieng2020topic,wang2020neural})
to extract the main concepts discussed in the psychotherapy. We use the Embedded Topic Model (ETM) \cite{wang2020neural} in this work because it was shown to create the most diverse concepts in psychological corpus \cite{lin2022neural}. 
% One can also adopt a symbolic approach to the topic modeling to gain further insights into the causalities and relationships between these topical concepts as in \cite{lin2022NSTM}. 
In this study, we use annotate each turn with their most likely topic and identifies seven unique topics (Topic 0 is about figuring out, self-discovery and reminiscence; Topic 1 is about play. Topic 2 is about anger, scare and sadness. Topic 3 is about counts. Topic 6 is about explicit ways to deal with stress, such as keep busying and reaching out for help. Topic 7 is about numbers. Topic 8 is about continuation and keep doing.)
% [0, 6, 7, 8, 1, 2, 3]

\textbf{Recommendation system setting.}
Then, we pair these ``items'' with the ``\textit{users}'' and ``contents'', which in our case, would be the patientID, his or her previous turns, their aggregated formats and other meta data. For instance, we know that within each sessions, there exists many pairs of turns, and they would belong to the same ``user''. However, one can also assign all turns belong to one clinical label, or all turns related to a certain topic as one ``user''. In this example, we choose the session ids as users. And lastly, the ``ratings'' would be patient's inferred alliance scores predictive of the therapeutic outcomes. Creating this database from historical data, we can train our system. Since we have defined our users, items, contents and ratings, the recommendation engine can be easily crafted with content-based
\cite{pazzani2007content,basu1998recommendation,aggarwal2016recommender} 
and collaborative filtering
\cite{sarwar2001item,he2017neural,koren2022advances,su2009survey}. 
% As the first step, we use the item-based collaborative filtering \cite{sarwar2001item} as our engine. 
Since our session turns are sequential and can specify a state or timestamp, it might be suitable for reinforcement learning (RL)
\cite{zheng2018drn,wang2014exploration,zou2020pseudo} 
and session-based approaches
\cite{li2017neural,wu2019session,ludewig2018evaluation}, which can be neuroscience or psychiatry-inspired \cite{lin2019split,lin2020story,lin2021models,lin2020unified} to provide further interpretable clinical insights.
During the deployment, our system registers our session as a new ``user'' if we adopt a session-based item.
% providing punctuated rater evaluations as inference anchors. Next steps include predicting these inference anchors as states (like \cite{lin2022neural,lin2020predicting}) and training chatbots as reinforcement learning agents given these states (like \cite{}).

\textbf{Deep reinforcement learning recommendation approaches.} Reinforcement learning approaches are effectively applied in language and speech tasks (as reviewed in 
\cite{lin2022rl4lang}), among which recommendation is an important use case. As shown in the right panel of Figure \ref{fig:pipeline}, the reinforcement learning environment is formulated such that the recommendation agent takes an action by recommending a strategy (say, a discussion topic). And the therapist will interact with the patient taking that suggestion into account. The dialogue interaction, in turn, has a quality evaluation of some sort (say, the therapeutic working alliance score). This serves as a reward to the recommendation agent to update its weights. In the meanwhile, the state is progressed to the next therapeutic states. 
As a first step, we evaluate three popular deep RL algorithms. Based on the deterministic policy gradient in an actor-critic architecture, the Deep Deterministic Policy Gradients (DDPG) \cite{lillicrap2015continuous} is a model-free algorithm for continous action spaces, and one of the first successful algorithms to learn policies end-to-end. Building upon the Double Q-Learning \cite{hasselt2010double}, Twin Delayed DDPG (TD3) \cite{fujimoto2018addressing} is a similar solution is proposed to correct for the overestimated value issue, and yields more competitive results in various game settings. As the online data collection of RL models are usually time consuming, in real world industrial setting, these models are usually trained using previously collected data. As a result, there is a growing popularity of offline reinforcement learning methods \cite{levine2020offline}. Among them, Batch Constrained Q-Learning (BCQ) \cite{fujimoto2019off} is the first continuous control deep RL algorithm that yields competitive results in off policy evaluations by restricting the agent's exploration in the action space.

\begin{table}[tb]
% \begin{minipage}{\linewidth}
      \caption{Pearson's r of the actual actions taken in the test set with their predicted actions
      }
      \label{tab:rs_eval} 
      \centering
      \resizebox{\linewidth}{!}{
     \begin{tabular}{l | c | c | c | c | c }
 & Anxi & Depr & Schi & Suic & All \\ \hline
R2D2-DDPG-TASK & \textbf{0.3796} & 0.3376 & 0.1556 & 0.3292 & 0.0578 \\ 
R2D2-DDPG-BOND & 0.2417 & 0.3838 & 0.1539 & 0.0873 & 0.1455 \\ 
R2D2-DDPG-GOAL & 0.0761 & 0.3682 & 0.4589 & -0.0210 & 0.2243 \\ \hline
R2D2-TD3-TASK & 0.0707 & 0.1310 & 0.0443 & 0.3188 & 0.3357 \\ 
R2D2-TD3-BOND & 0.2018 & 0.3363 & 0.0908 & 0.3070 & 0.1101 \\ 
R2D2-TD3-GOAL & 0.0984 & 0.2222 & \textbf{0.4599} & 0.2044 & \textbf{0.3765} \\ \hline
R2D2-BCQ-TASK & 0.1128 & \textbf{0.4042} & 0.1401 & 0.1422 & 0.0825 \\ 
R2D2-BCQ-BOND & 0.0778 & 0.0876 & 0.0987 & \textbf{0.4152} & 0.0885 \\ 
R2D2-BCQ-GOAL & 0.0810 & 0.1231 & 0.0833 & 0.0788 & 0.0780 \\ 
\end{tabular}
 }
%  \end{minipage}
% \vspace{-1em}
\end{table}

\textbf{Reinforced Recommendation model for Dialogue topics in psychiatric Disorders (R2D2).}
Combining all the elements, we have our R2D2 model (Figure \ref{fig:pipeline}B). Here we identify each session as a user, and the states are frames of dialogues which can be labelled their topics in real time and their ratings with a working alliance (WA) inference module. The reinforcement learning core, powered by deep RL, predicts the best action represented by an embedding for the items (topics). This embedding can be pre-computed, for instance, using dimension reduction techniques to find clusters of different topics in a low-dimensional space. We use the Doc2Vec embedding of the original dialogue turns, averaged by their topic labels, such that each action (i.e. the topic id) have an averaged representation in the sentence embedding space. This action representation can be translated into a topic label with nearest neighbor, and a given dialogue response will be selected from the historical dialogue data to continue the conversation. The reward can then be computed using the working alliance rate.

\section{Empirical results}

% \begin{figure}[tb]
% \centering
%     \includegraphics[width=\linewidth]{Figures/results.png}
% 
% \caption{Empirical results on MiniVox and AlexStreet datasets
% }\label{fig:results}
% \end{figure}

%  \begin{figure*}[t]
% \centering
% \begin{tabular}{c | c | c | l}
%     \begin{tabular}{c }
%     {\includegraphics[width=.32\linewidth]{Figures/ddpg_loss.png}} \end{tabular}
%     &
%     \begin{tabular}{c }
%     {\includegraphics[width=.16\linewidth]{Figures/td3_loss_up.png}} \\ {\includegraphics[width=.16\linewidth]{Figures/td3_loss.png}}  \end{tabular}
%     &
%     \begin{tabular}{c}
%     {\includegraphics[width=.16\linewidth]{Figures/bcq_loss_up.png}} \\ {\includegraphics[width=.16\linewidth]{Figures/bcq_loss.png}}  \end{tabular}
%     &
%     \begin{tabular}{l | c }
% Agent & r \\ \hline
% DDPG & 0.2712 \\
% TD3 &  0.2192 \\
% BCQ & 0.2843
% \end{tabular}
% \end{tabular}
% \caption{Evaluation of deep reinforcement learning recommenders. (Left three) loss functions for the sub-networks for DDPG, TD3, and BCQ. (Right) Pearson's r of the actual actions taken in the test set with their predicted actions.
% }\label{fig:rs_eval}
% \end{figure*}

 \begin{figure*}[t]
\centering
\begin{tabular}{l | r}
    % \fbox{\includegraphics[width=.43\linewidth]{Figures/step1.png}}
    {\includegraphics[width=.44\linewidth]{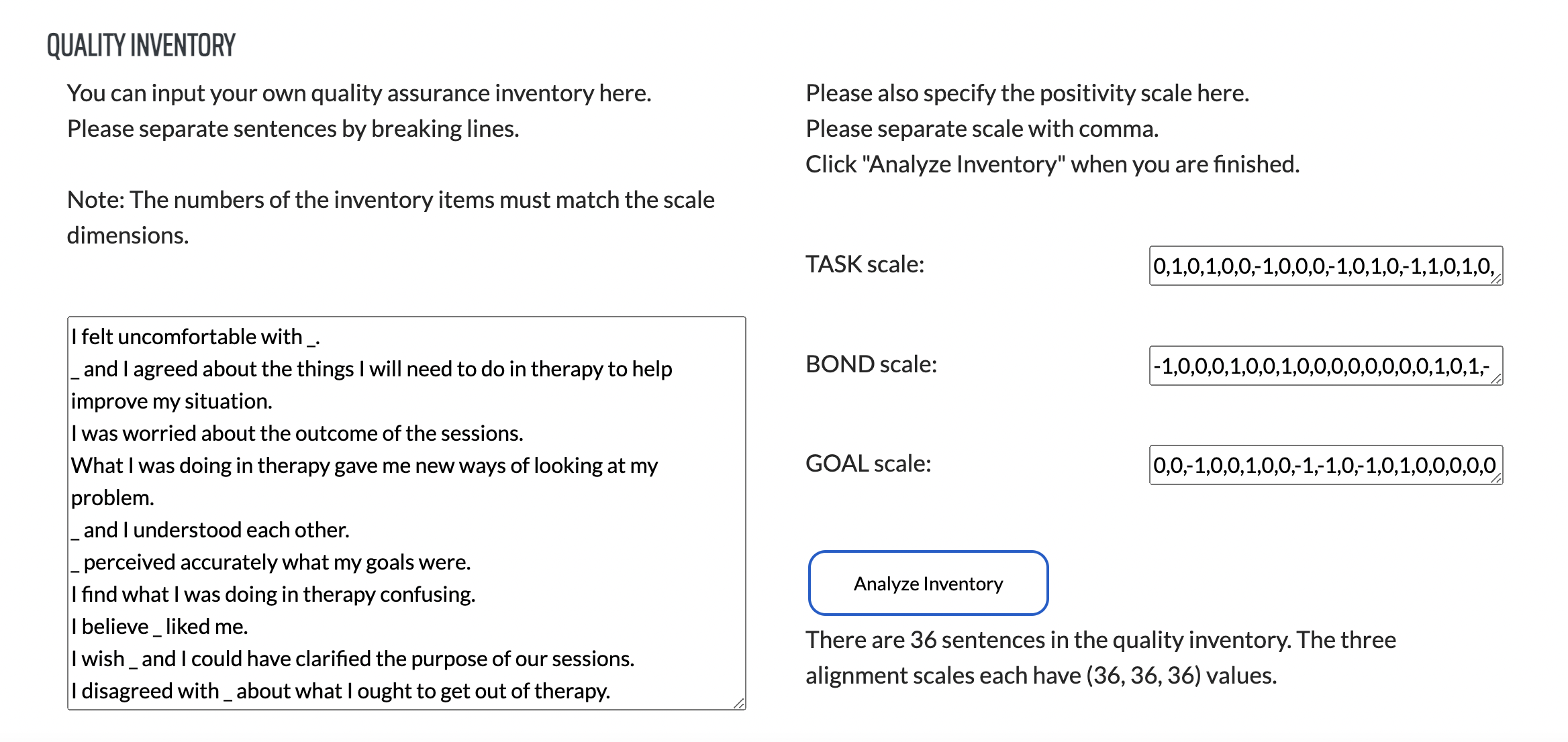}} &
    {\includegraphics[width=.48\linewidth]{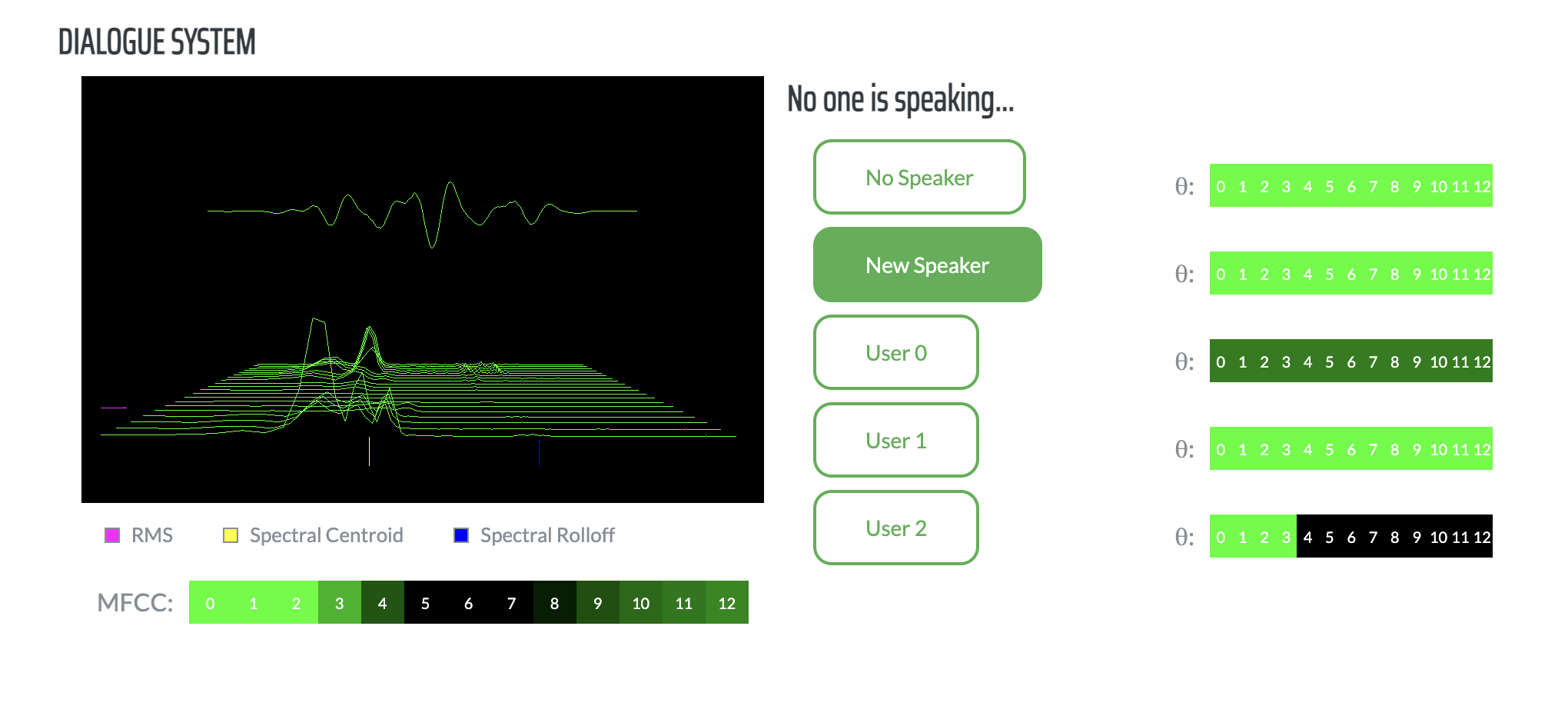}} \\ \hline
    {\includegraphics[width=.44\linewidth]{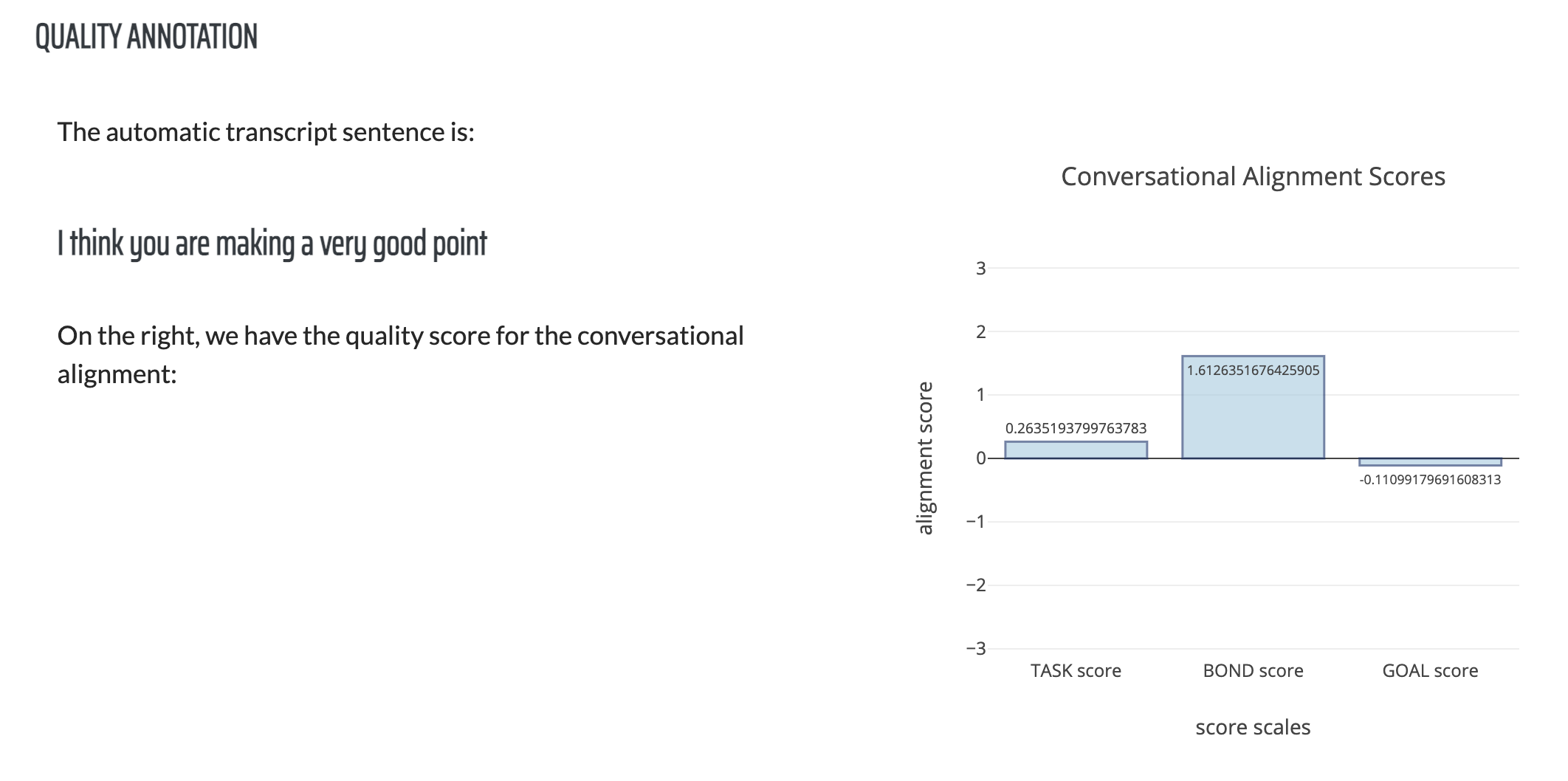}} &
    {\includegraphics[width=.48\linewidth]{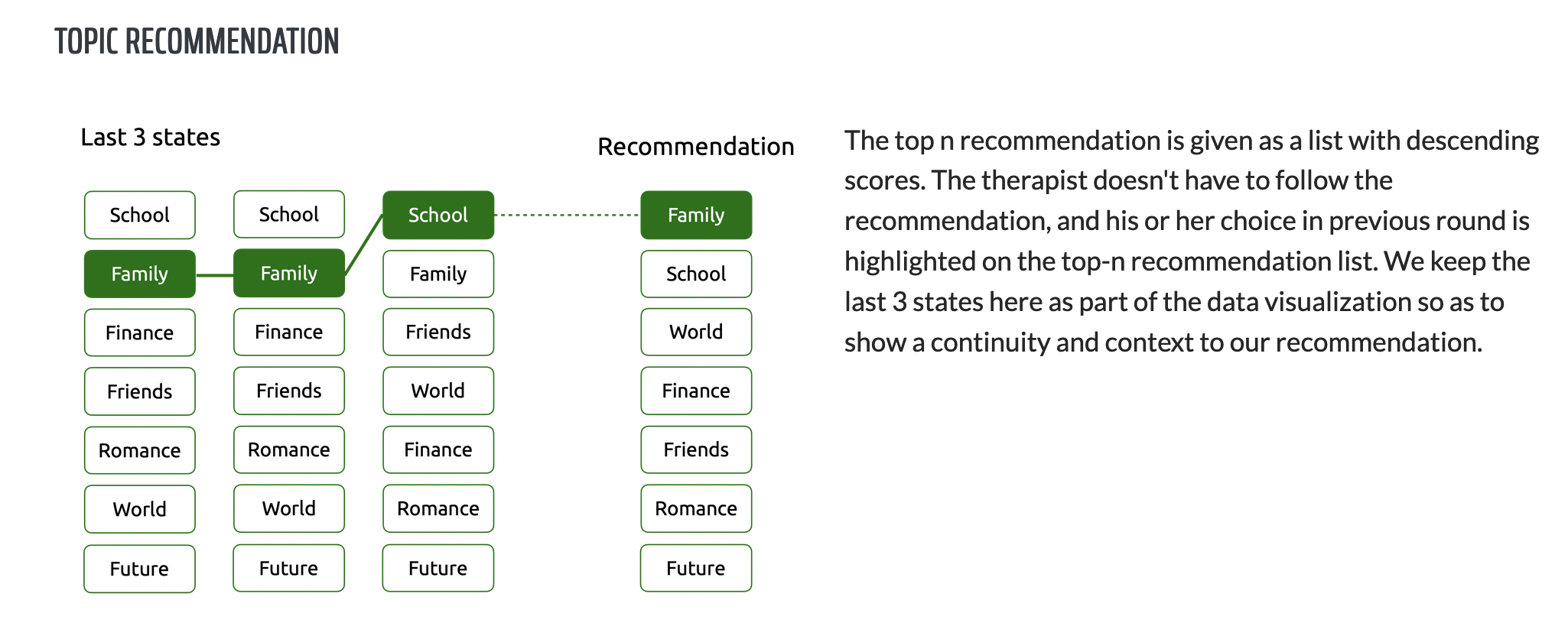}}
\end{tabular}
\caption{State screenshots of SupervisorBot web app: inventory inputs, diarization,  annotation, strategy recommendation.
}\label{fig:sbot}

\end{figure*}

\textbf{Experimental setting.}
To evaluate the recommendation systems, we preprocess the Alex Street psychotherapy dataset \footnote{https://alexanderstreet.com/products/counseling-and-psychotherapy-transcripts-series}, which consists of transcribed recordings of over 950 therapy sessions between multiple anonymized therapists and patients, into a recommendation system format (219,999 recommendation actions) and then split it into 95/5 train-test sets. The dataset consists of four types of psychiatric conditions: anxiety, depression, schizophrenia and suicidal cases. We train R2D2 on each of the text corpus, as well as on all four together. To set up the batch training for reinforcement learning, we cut the turns into frames of 10 turn pairs and use a batch size of 32. We represent the action spaces (the topics to recommend) in three candidate embedding spaces: the averaged 300-dimension Doc2Vec embedding for each topic, the averaged 36-dimension principal component analysis (PCA) embedding, and the averaged 2-dimension Uniform Manifold Approximation and Projection (UMAP) embedding. Due to the space limit, we only present the results for the first embedding, but leave the others in appendix. We train the R2D2 with three reinforcement learning agents (DDPG, TD3 and BCQ) each for 50 epochs, where their losses consistently drop and converge in a stable way. Based on the loss curve, there are no overfitting in all model training processes. 

\textbf{Empirical results.}
To evaluate the performance of the three recommendation agents, we compute the Pearson's r of the recommended actions with their corresponding ground truth actions the test set (Table \ref{tab:rs_eval}). Since we are the first system in this application problem, there are no state-of-the-art or baseline so far. Instead, we compare among variants of R2D2. Other than testing on different subset of the datasets and reinforcement learning algorithms, we also use three different scales of working alliance as our ratings: task, bond and goal, which measures different aspects of emotional alignments in psychotherapy. 
We observe that the best performing model for four disorders are: R2D2-DDPG-TASK for depression sessions with a correlation of 0.3796, R2D2-BCQ-TASK for depression session (0.4042), R2D2-TD3-GOAL for schizophrenia sessions (0.4599) and R2D2-BCQ-BOND for suicidal sessions (0.4152). If we consider all four classes together, R2D2-TD3-GOAL appears to be the best performing models (0.3765). We notice that the DDPG and TD3 bases of R2D2 yields similar rankings among using three working alliance scales as their ratings, while the BCQ tends not to. For instance, in schizophrenia cases, the alignment in the goal scale appear to provide a far more advantageous recommendation prediction than the other two implicit feedbacks (task and bond alignments), while in R2D2-BCQ, the effect is less pronounced. For specific disorders, R2D2-DDPG is the recommender winner for anxiety, depression and schizophrenia, and R2D2-TD3 is the winner for suicidal cases (which should be taken with a grain of salt considering the small amount of data we have on them). When pooling the sessions of four disorders together, the recommender winner appears to be R2D2-TD3, which may suggest that R2D2-TD3, given its twin delayed mechanism to correct for value overestimation, are better suited for heterogeneous rather than homogeneous cases. It was a surprise that R2D2-BCQ doesn't demonstrate in our dataset, an advantage to constrain the possible extrapolation errors by the non-offline methods. This evaluation provides a proof of concept. Future work will focus on systematically comparing a larger spectrum of deep reinforcement learning and model architectures.

% More in \cite{lin2022deep}.
% We can be generalize in a sense that with this approach one can go over your sessions (as a therapist) and analyze the dynamics afterwards.

\textbf{Ethical considerations.} Following the ethical guidelines 
% in \cite{matthews2017stories,graham2019artificial} 
and the operational suggestions in \cite{lin2022ethics}, we make sure that all training examples that we evaluate on are properly anonymized with pre- and post-processing techniques, and disclaim that these investigations are proof of concept and require extensive caution to prevent from the pitfall of over-interpretation.

\section{Web-Based System: SupervisorBot}

``SupervisorBot'' is an interactive web-based system (Fig \ref{fig:sbot}). We first give users the instructions on how to use the system. Then they are lead to input their own inventory used to evaluate the dialogue quality. In this case, we put in a default one, using the working alliance inventory. They are guided to input the score scale corresponding to each inventory item and click on ``Analyze'' to finalize. In the speaker diarization part, we compute and visualize the Mel Frequency Cepstral Coefficients (MFCC) in a sliding window fashion given the real-time audio input from microphone, with the MFCC bands color coded in the page. 
% Finishing these two steps as the preparation, the system is now running, and the therapist can sit back and go on with the session. 
% At the start, there are only two buttons available:  ``No Speaker'' and ``New Speaker''. The agent chooses an arm by setting it be highlighted. If it is correct, we do not have to change it (unless it's ``New Speaker'', where we need to click on it to confirm creating a new arm). The feature band $\hat{\mathbf{\theta}}_a$ of each arm is also color coded real-time to visualize how the agent learns across trials. If it is incorrect, we click on the right arm to give the system a feedback.
The app now moves to the annotation panel, where the therapist can see that a transcript is displayed, along with who is speaking. The computed alliance score in the three scales are also dynamically displayed in real-time according to the content of the dialogue turn. This is helpful information to assist the therapist. And in our last panel, we have our recommendation guidance. The topics to choose from are ranked and top N are displayed. The therapist can use it as a hint and initiate his response given a top recommendation. The system will transcribe his response and highlight the topic he most likely ended up choosing in the last round, and save that information as part of historical data. The system refreshes its parameters at the end of each session.

\section{Conclusions and Future Directions}

In this work, we provide a practical example of how a real-time recommendation system can help therapists better treat their patients in psychotherapy sessions with informative clinical annotations and recommendations of treatment strategies with deep reinforcement learning. Although in this example, the strategies are the topics for the therapist to initiate or continue, the same approach can be extended to more complex and nuanced treatment suggestions. For instance, in the ABC approach of cognitive behavioral therapy (CBT), our system can suggest a belief (B) to guide the patients to better understand the causality between the activating event (A) and its consequence (C). 

Before we conclude, another interesting perspective to view this line of research is hidden in Figure \ref{fig:pipeline}: while the recommendation agent is driven by reinforcement learning, the therapist (and even patient) have their agency which updates under the reinforcement learning principles. For instance, the patient can directly offer feedback to the therapists. And given the feedback, the therapist may adjust his or her internal model to weigh on the quality of the suggestions by the recommendation agent. Next steps include modeling these theory of minds and confidence levels in this multi-participant human computer interaction setting.
% providing punctuated rater evaluations as inference anchors. Next steps include predicting these inference anchors as states (like \cite{lin2022neural,lin2022predicting}) and training chatbots as reinforcement learning agents given these states (like \cite{lin2020story,lin2021models,lin2020unified}).

% \section{Resources}

% \noindent
% \textbf{Website:} \href{https://www.baihan.nyc/viz/SupervisorBot}{https://www.baihan.nyc/viz/SupervisorBot}

% \noindent
% \textbf{Codes}: \href{https://github.com/doerlbh/PsychiatryNLP}{https://github.com/doerlbh/PsychiatryNLP}

% \vfill\pagebreak

% \section{REFERENCES}
% \label{sec:refs}

% \newpage
\bibliographystyle{IEEEbib}
\bibliography{main}

\end{document}